\documentclass[runningheads]{llncs}
\usepackage{graphicx}
\usepackage{amsmath}
\usepackage{amssymb}
\usepackage{algorithm}
\usepackage{algorithmic}

\begin{document}
\title{Improving Randomized Learning of Feedforward Neural Networks by Appropriate Generation of Random Parameters\thanks{Supported by Grant 2017/27/B/ST6/01804 from the National Science Centre, Poland.}}
\titlerunning{Improving Randomized Learning of Feedforward Neural Networks ...}
\author{Grzegorz Dudek}%\inst{1}\orcidID{0000-0002-2285-0327}}
\authorrunning{G. Dudek}
\institute{Electrical Engineering Department, Czestochowa University of Technology, Czestochowa, Poland\\	
	\email{dudek@el.pcz.czest.pl}}
\maketitle              % typeset the header of the contribution
\begin{abstract}
In this work, a method of random parameters generation for randomized learning of a single-hidden-layer feedforward neural network is proposed. The method firstly, randomly selects the slope angles of the hidden neurons activation functions from an interval adjusted to the target function, then randomly rotates the activation functions, and finally distributes them across the input space. For complex target functions the proposed method gives better results than the approach commonly used in practice, where the random parameters are selected from the fixed interval. This is because it introduces the steepest fragments of the activation functions into the input hypercube, avoiding their saturation fragments.

\keywords{Function approximation \and Feedforward neural networks \and Neural networks with random hidden nodes \and Randomized learning algorithms.}
\end{abstract}

\section{Introduction}

Feedforward neural networks (FNNs) learn from data by iteratively tuning their parameters, weights and biases, using some form of gradient descent method. Due to the layered structure of the network, the learning process is complicated, inefficient and time consuming. The network converges to the local optima, and the final result is very sensitive to the initial values of parameters.

Some of these drawbacks can be avoided by using a randomized learning approach. In this case the weights and biases of the hidden nodes are randomly selected from given intervals according to any continuous sampling distribution and remain fixed. The only parameters that are learned are the weights between the hidden layer and the output layer. The resulting optimization task becomes convex and can be formulated as a linear least-squares problem \cite{Pri15}. So, the problem can be considered as a linear one and the gradient descent method is not needed to solve it. The output weights can be analytically determined through a simple generalized inverse operation of the hidden layer output matrices. Due to the non-iterative nature, the randomized learning of FNNs can be much faster than classical gradient descent-based learning. In addition, it is easy to implement in any computing environment.

It was theoretically proven that FNN is a universal approximator for a continuous function on a bounded finite dimensional set, when the random parameters are selected from a uniform distribution within a proper range \cite{Igi95}. Husmeier showed that the universal approximation property also holds for a symmetric interval for random parameters if the function to be approximated meets the Lipschitz condition \cite{Hau99}. However, how to select the range for the random parameters remains an open question. It is well known that this range has a significant impact on the performance of the network. This has already been noted in early works on randomized learning algorithms, e.g. authors of \cite{Pao94} and \cite{Hau99} recommended optimizing the interval for a specified task. In \cite{Li17} a series of simulations were carried out to illustrate the significance of the range of random parameters on modeling performance. The authors empirically showed that a widely used setting for this range, usually $[-1, 1]$, is misleading because the network is unable to model nonlinear maps, no matter how many training samples are provided or what sized networks are used. Although, they observed that for some specific ranges the network performs better in both learning and generalization than for other ranges, they do not give any tips on how to select an appropriate range. 

In \cite{Zha16c} the problem of the random parameters range is investigated by introducing a scaling factor to control the ranges of the randomization: $[-s, s]$ is used for weights and $[0,s]$ for biases. The authors observed that the commonly adopted approach where $s = 1$ may not lead to optimal performance. The network performs poorly when the range of the random parameters becomes either too large or too small. Setting small $s$ to increase the discrimination power of the features in the hidden neurons may cause more neurons to saturate. On the other hand, setting large $s$ to reduce the possibility of neuronal saturation may reduce the discrimination power of the features in the hidden neurons. To find the optimal symmetric interval for weights and biases in \cite{Wan17} a stochastic configuration algorithm is used. Random parameters are generated with an inequality constraint adaptively selecting the range for them, ensuring the universal approximation property of the model.

In \cite{Gor16} it was noted that if the network nodes are chosen at random and not subsequently trained, they are usually not located in accordance with the density of the input data. Consequently, the training of linear parameters becomes ineffective at reducing errors in the nonlinear part of the network. Moreover, the number of nodes needed to approximate a nonlinear map grows exponentially, and the model is very sensitive to the random parameters. To improve the effectiveness of the network, the unsupervised placement of network nodes according to the input data density could be combined with the subsequent supervised or reinforcement learning values of the linear parameters of the approximator.

Despite intensive research in recent years into randomized learning of FNNs, there are still several open problems which need to be addressed, such as how to generate random parameters. This issue remains untouched in the literature and is considered to be one of the most important research gaps in the field of randomized algorithms for training NNs \cite{Zha16s}, \cite{Cao18}. In many practical applications in classification or regression problems the ranges are selected as fixed without scientific justification, typically $[-1, 1]$, regardless of the data and activation function type. 

Recently, in \cite{Dud19} a new method of random parameters generation was proposed. The formulas for weights and biases were derived assuming that the steepest fragments of the activation functions are located in the input space region and their slopes are adjusted to the target function complexity. This method of generating random parameters allows us to control the generalization degree of the model and leads to an improvement in the approximation performance of the network.    

This work follows on from \cite{Dud19}. Here we propose an alternative way of initially setting the activation functions for single-hidden-layer FNN, where we randomly determine their slope angles and rotations in space. Instead of selecting the weights and biases from the assumed interval, we randomly select the slope angles from the interval adjusted to the target function, and then randomly rotating the activation functions and shifting them into the input hypercube, we calculate weights and biases. As shown in Section 3, the proposed method gives much better results than the standard approach with fixed intervals for random parameters. It is also more intuitive than the method proposed in \cite{Dud19} as it has parameters which are easily interpreted.

\section{Generating Random Parameters of Hidden Nodes}
\label{GHN}
For brevity, we use the following acronyms:    
\begin{description}
	\item CSs - constructional sigmoids, i.e. the set of sigmoid activation functions of hidden nodes whose linear combination forms the function fitting data,
	\item II - input interval,
	\item FF - fitted function (curve or surface) constructed by FNN,
	\item TF - target function.
\end{description}

In this section we analyze how the random parameters affect the approximation abilities of FNN. The standard approach for generating random parameters is analyzed and a new approach is proposed. We consider a single-hidden-layer FNN with one output, $m$ hidden neurons and $n$ inputs. A sigmoid is used as an activation function. The output weights are calculated using the Moore-Penrose pseudo-inverse operation. 

To illustrate the results, let us use a single-variable TF in the form:      
\begin{equation}
g(x) = \sin\left(20\cdot\exp(x)\right)\cdot x^2
\label{eqTF1}
\end{equation}
where $ x \in [0, 1] $.

\begin{figure}
	\centering
	\includegraphics[width=0.49\textwidth]{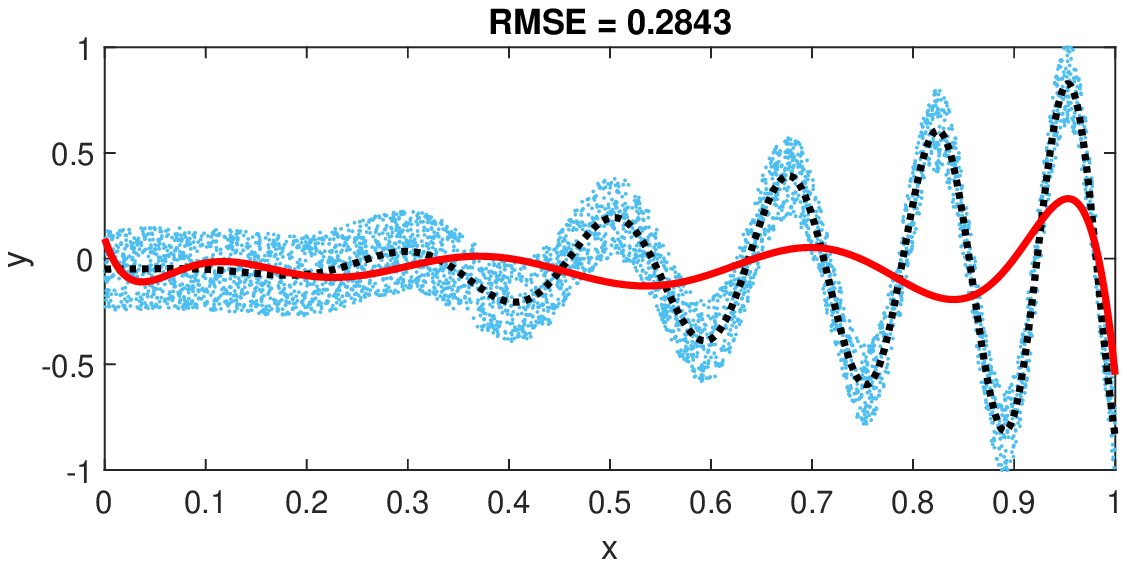}
	\includegraphics[width=0.49\textwidth]{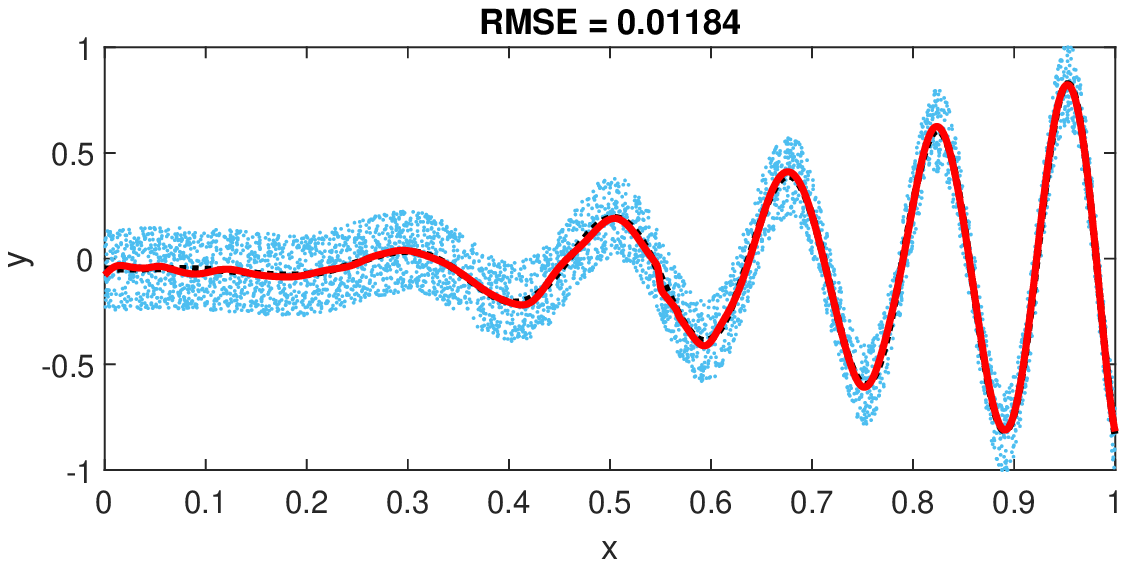}
	\includegraphics[width=0.49\textwidth]{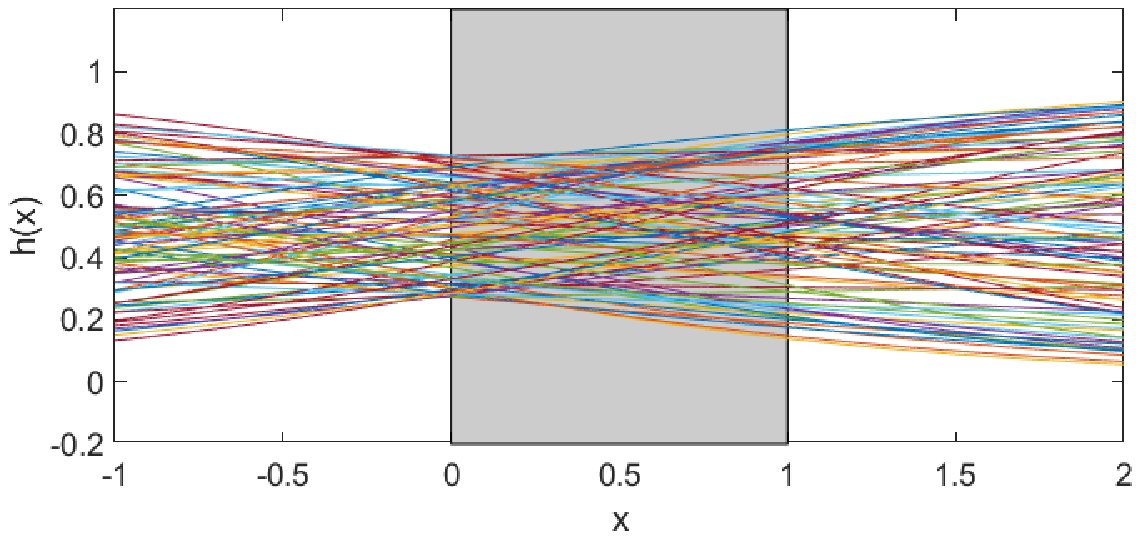}
	\includegraphics[width=0.49\textwidth]{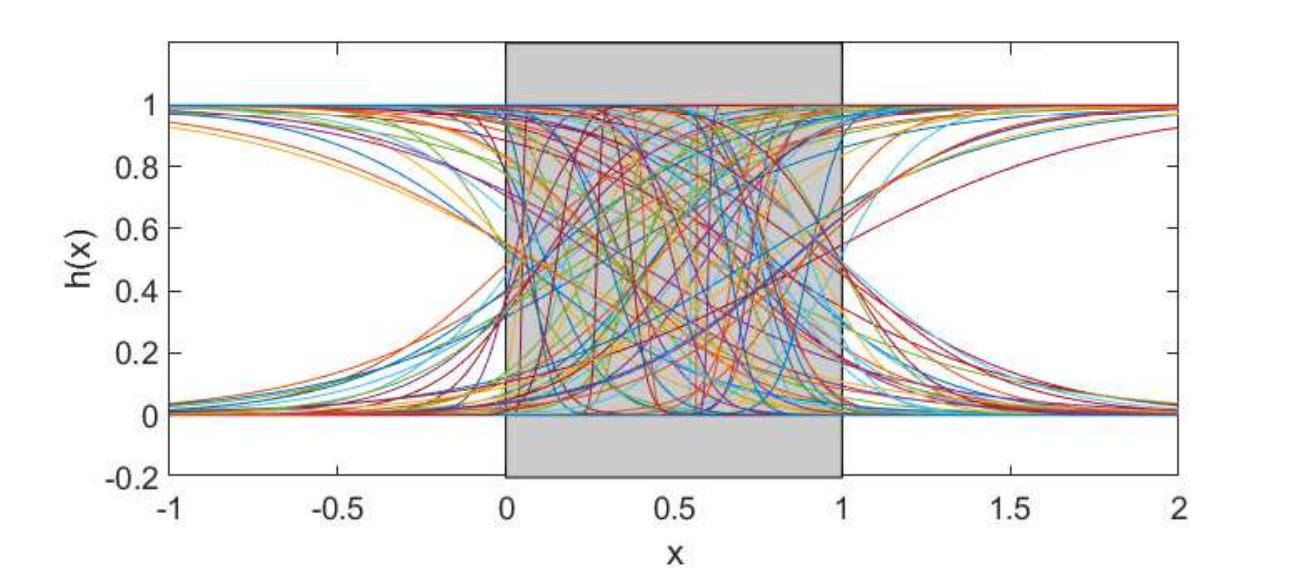}
	\includegraphics[width=0.49\textwidth]{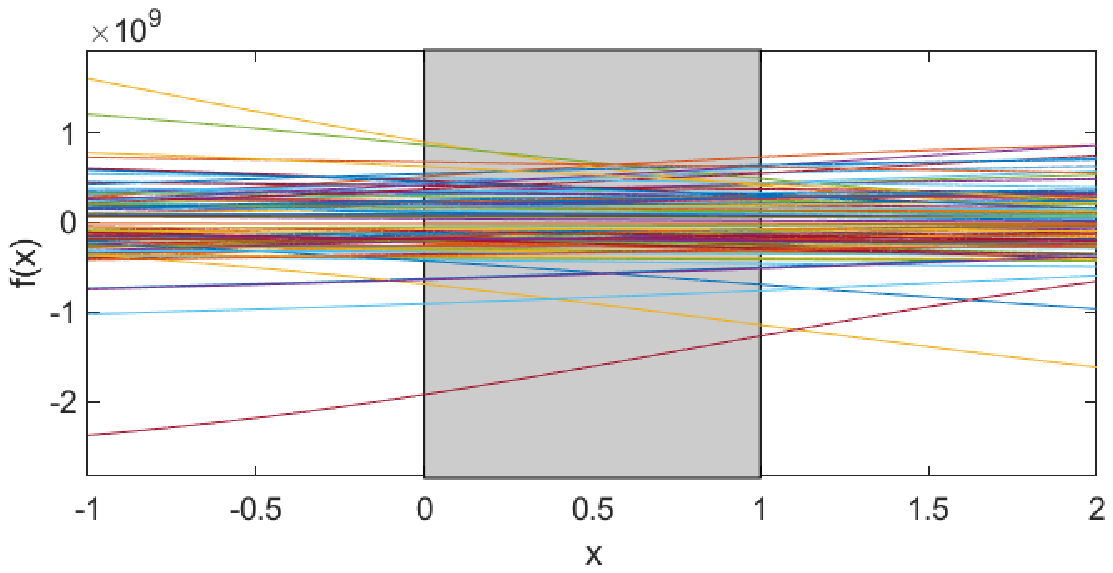}
	\includegraphics[width=0.49\textwidth]{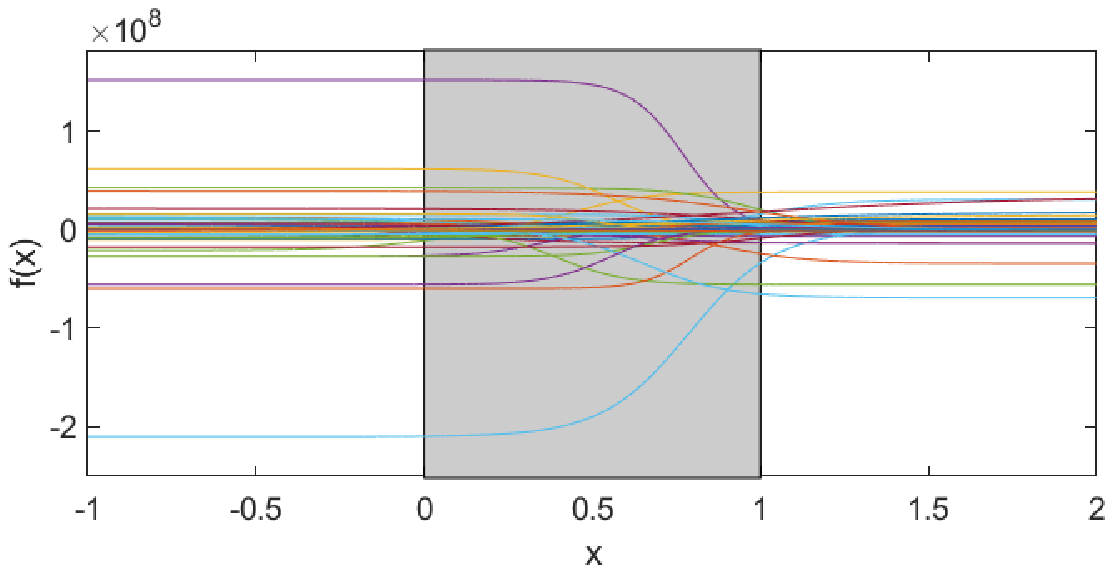}
	\caption{Fitted curve (upper panel), CSs (middle panel) and weighted CSs (bottom panel) for the standard method, $a, b \in [-1, 1]$ (left panel), and the proposed method, $\alpha_{min}=30^\circ$ (right panel).} 
	\label{Fig1}
\end{figure}

This function is shown by the dashed line in the upper chart of Fig. \ref{Fig1}. As can be seen in this figure, variation of TF (\ref{eqTF1}) increases along the II =$ [0, 1] $. At the left bound of the II the TF is flat, while towards the right bound its fluctuations increase.

For NN learning we generate a training set containing $ 5000 $ points $ (x_l, y_l) $, where $ x_l $ are uniformly randomly distributed on $ [0, 1] $ and $ y_l $  are calculated from \eqref{eqTF1} and then distorted by adding the uniform noise distributed in $ [-0.2, 0.2] $. A test set of the same size is created similarly but without noise. The outputs are normalized in the range $ [-1, 1] $.

We consider the sigmoid activation function for hidden nodes: 

\begin{equation}
h(x) = \frac{1}{1 + \exp(-(a\cdot x + b))}
\label{eqSig}
\end{equation}
where $ a $ is a weight deciding about the slope of the sigmoid $ (\mathrm{d}h/\mathrm{d}x) $ and $ b $ is a bias shifting it along the x-axis. 

The set of sigmoids included in the hidden neurons are the basis functions whose linear combination forms the function fitting data. For nonlinear TF this set should deliver the nonlinear parts of sigmoids (avoiding their saturated fragments) to model the TF with required accuracy. The sigmoids should also be distributed properly in the II so that their steep fragments correspond to the steep fragments of the TF. Are these requirements met when the sigmoids weights and biases are both generated randomly from the typical interval $[-1, 1]$? 

The left panel of Fig. \ref{Fig1} shows results of function (\ref{eqTF1}) fitting when using FNN with 100 hidden nodes whose parameters $a$ and $b$ are both selected randomly from $[-1, 1]$. The upper chart shows the training points and the FF (solid line). The CSs are shown in the middle chart and CSs multiplied by the output weights are shown in the bottom chart. Note that CSs are flat in the II, which is shown as a gray field, and many of them have their steepest parts, which are around their inflection points (sigmoid value for the inflection point is 0.5), outside the II. Thus, the CSs slopes and their distribution in II do not correspond to TF fluctuations. This results in a very weak fitting. This simple example leads to the conclusion that random parameters of hidden nodes cannot be generated from the fixed interval $[-1, 1]$. The intervals for them should, instead, be estimated taking into account the II and the TF features, such as fluctuations.

\subsection{The Idea of the Proposed Method}

Instead of determining the interval for weights $a$, we determine the interval for the slope angles of CSs. By the slope angle $\alpha$ we mean the angle between a tangent line to the sigmoid at its inflection point and the x-axis. Using the slope angles instead of the weights is more intuitive because we can imagine without any effort a sigmoid having slope angle $\alpha$, but it is hard to imagine a sigmoid having the weight $a$. So, it is clear to us what CSs for $\alpha\in[30^\circ, 60^\circ]$ look like, and we have no idea what CSs for $a\in[2.31, 6.93]$ look like. In fact, they look similar because these are corresponding intervals.

The weight $a$ translates nonlinearly into the slope angle. To find the relationship between them, it should be noted that the derivative of a sigmoid at the inflection point is equal to $\tan\alpha$:
\begin{equation}
ah(x)(1-h(x)) = \tan\alpha
\label{eqDer1}
\end{equation}

Setting this sigmoid in such a way that its inflection point is in $x = 0$ we get: 

\begin{equation}
a\frac{1}{1 + \exp(-(a\cdot 0 + 0))} \left(1-\frac{1}{1 + \exp(-(a\cdot 0 + 0))}\right) = \tan\alpha
\label{eqDer2}
\end{equation}
From \eqref{eqDer2} we achieve:
\begin{equation}
a = 4\tan\alpha
\label{eqA}
\end{equation}

Note that for a typical interval for $a$, i.e. $[-1,1]$ we get $\alpha \in [-14^\circ, 14^\circ]$, so rather flat CSs. Note also that selecting $a$ uniformly from a certain interval (typical case) we get nonlinearly distributed angles $\alpha$, according to the tangent function. We propose in this work instead of selecting weights $a$ for CSs, randomly selecting the slope angles $\alpha$ for them from the interval:

\begin{equation}
\Gamma = (-90^\circ, -\alpha_{min}) \cup (\alpha_{min}, 90^\circ)
\label{eqInt1}
\end{equation}
where $\alpha_{min}\in[0^\circ, 90^\circ]$ is a limit angle adjusted to the TF. For lower value of $\alpha_{min}$ we have both flat and steep sigmoids in the set of CSs. When increasing $\alpha_{min}$ we generate more steep sigmoids instead of flat ones. Intuitively, functions with strong fluctuations require larger angle $\alpha_{min}$.

Having randomly selected slope angles $\alpha$ for CSs we can calculate weights $a$ for them from \eqref{eqA}. Now we can focus on the biases $b$, which decide about the sigmoids shifts. We would like to have the steep fragments of CSs inside the II. So, let us set CSs inflection points inside the II at the points $x^*$ which are randomly selected from the II. For the $i$-th sigmoid from the CSs set and any point $x_i^*$ selected for it from the II we obtain:
\begin{equation}
\frac{1}{1 + \exp(-(a_ix_i^* + b_i))}=0.5
\label{eqSig2}
\end{equation}

After transformations we obtain:
\begin{equation}
b_i=-a_ix_i^*
\label{eqbi}
\end{equation}

So the bias of the sigmoid strictly depends on its weight. Selecting $x_i^*$ uniformly from the II we get uniformly distributed CSs in the II. 

The right panel of Fig. \ref{Fig1} shows the results of function \eqref{eqTF1} fitting when using the proposed method to determine the weights and biases. The limit slope angle was selected as $\alpha_{min} = 30^\circ$. This means that the weights are from the interval $(-\infty, -2.31) \cup (2.31, \infty)$. But note that we do not uniformly select weights from this interval. If that were the case, then most of them would have high values corresponding to angles close to $90^\circ$. Instead, we select the angles randomly from the uniform distribution on \eqref{eqInt1}, and then calculate weights. Compare the right panel of Fig. \ref{Fig1} with the left one and note the completely different CSs distribution and slopes resulting in a pretty good fit.

\subsection{Multidimensional Case}

\begin{figure}
	\centering
	\includegraphics[width=0.49\textwidth]{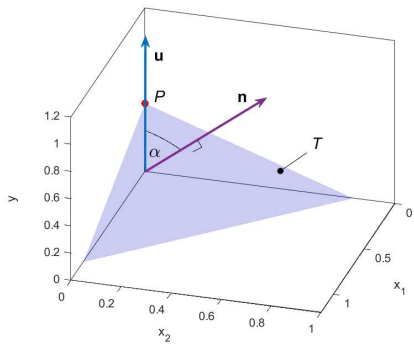}
	\caption{Slope angle $\alpha$ as an angle between the normal vector $\mathbf{n}$ and the unit vector $\mathbf{u}$.} 
	\label{Fig3}
\end{figure}

The idea of random parameters generation is now expanded for the multi-dimensional case. Let us consider a multi-dimensional sigmoid $S$ of the form:
\begin{equation}
h(\mathbf{x}) = \frac{1}{1 + \exp\left(-\left(\mathbf{a}^T\mathbf{x} + b\right)\right)}
\label{eqSigM}
\end{equation}
which has one of its inflection points, point $P$, in the origin of the Cartesian coordinate system, i.e. in  $O=(0, 0, ..., 0)$. (The sigmoid inflection points are all points for which $h(\mathbf{x})=0.5$). Let $T$ be the tangent hyperplane to the sigmoid at point $P$. This hyperplane takes the form:
\begin{equation}
a_1^{\prime}x_1+...+a_n^{\prime}x_n+a_0^{\prime}y+b^{\prime} = 0
\label{eqPl}
\end{equation}
where $x_i$ are $n$ independent variables and $y$ is a dependent variable. A normal vector to this hyperplane is $\mathbf{n}=[a_1^{\prime}, ..., a_n^{\prime}, a_0^{\prime}]^T$. Let $\alpha \in (0^\circ, 90^\circ)$ be an angle between the normal vector $\mathbf{n}$ and the unit vector in the direction of the y-axis: $\mathbf{u}=[0, ..., 0, 1]^T$ (see Fig. \ref{Fig3}). The vector $\mathbf{u}$ is normal for a hyperplane containing all x-axes. So, $\alpha$ is also an angle between this hyperplane and $T$, and decides about the slope of the sigmoid. When $\alpha$ is near $0^\circ$ we get flat sigmoid; when $\alpha$ is near $90^\circ$ we get very steep sigmoid. 

The cosine of this angle is expressed as:

\begin{equation}
\cos\alpha =\frac{\mathbf{u}\cdot\mathbf{n}}{\|\mathbf{u}\|\|\mathbf{n}\|}=\frac{0\cdot a_1^{\prime}+...+0\cdot a_n^{\prime}+1\cdot a_0^{\prime}}{1\cdot\sqrt{(a_1^{\prime})^2+...+(a_n^{\prime})^2+(a_0^{\prime})^2}}=\frac{a_0^{\prime}}{\sqrt{(a_0^{\prime})^2+...+(a_n^{\prime})^2}}
\label{eqCos1}
\end{equation} 
  
From this equation we obtain:

\begin{equation}
a_0^{\prime} = \pm \sqrt{\frac{\cos^2 \alpha}{1-\cos^2 \alpha} \left( (a_1^{\prime})^2+...+(a_n^{\prime})^2 \right)}= \pm \frac{\sqrt {(a_1^{\prime})^2+...+(a_n^{\prime})^2}}{\tan \alpha}
\label{eqa0}
\end{equation} 
where $\cos \alpha \neq 1$, $\tan \alpha \neq 0$, i.e. $\alpha \neq 0$.

Now we can construct the hyperplane $T$ inclined to the hyperplane containing all x-axes at an angle of $\alpha$, passing through point $P$ and randomly rotated around the y-axis. To do so, first we assume $\alpha\in(0^\circ, 90^\circ)$. Then, we generate randomly the first $n$ components of the normal vector $\mathbf{n}$: $a_1^{\prime}, ..., a_n^{\prime}$. Finally, we calculate its last component from $\eqref{eqa0}$. Note that this hyperplane defines the random rotation of the sigmoid around the y-axis and also its slope angle, which is $\alpha$. It is convenient to express hyperplane $T$ in the form:
\begin{equation}
y=-\frac{a_1^{\prime}}{a_0^{\prime}}x_1-...-\frac{a_n^{\prime}}{a_0^{\prime}}x_n+0.5
\label{eqPl2}
\end{equation}

We achieve this equation from \eqref{eqPl} assuming that $T$ passes through $P=(0,...,0,0.5)$, so the intercept term must be $0.5$.

Having randomly rotated a tangent hyperplane to sigmoid $S$, we are looking for weights $a_k$. A partial derivative of $S$ with respect to $x_k$ is:

\begin{equation} 
\begin{split}
\frac{\partial h(\mathbf{x})}{\partial x_k} &= a_k h(\mathbf{x})(1-h(\mathbf{x}))\\
&= a_k \frac{1}{1 + \exp(-(\mathbf{a}^T\mathbf{x} + b))}\left(1-\frac{1}{1 + \exp(-(\mathbf{a}^T\mathbf{x} + b))}\right)
\end{split}
\label{eqDer3}
\end{equation}

Sigmoid $S$ passes through point $P$, so $b$ must be $0$. Derivative \eqref{eqDer3} in $P$, where $x_1, ..., x_n=0$ and $b=0$, is:

\begin{equation} 
\frac{\partial h(P)}{\partial x_k} = \frac{1}{4} a_k
\label{eqDer4}
\end{equation}  

A partial derivative of hyperplane \eqref{eqPl2} with respect to $x_k$ is $-a_k^{\prime}/a_0^{\prime}$. Because $T$ is tangent to sigmoid $S$ in $P$, their derivatives in $P$ are the same, so: $1/4\cdot a_k = -a_k^{\prime}/a_0^{\prime}$, and finally:  

\begin{equation} 
a_k = -4\frac{a_k^{\prime}}{a_0^{\prime}}
\label{eqai}
\end{equation}   

This equation expresses the relationship between the sigmoid weights and the normal vector to the tangent hyperplane defining the sigmoid slope and rotation. We propose to select the slope angles for CSs from the interval:

\begin{equation}
\Delta = (\alpha_{min}, 90^\circ)
\label{eqInt2}
\end{equation}
where $\alpha_{min}\in[0^\circ, 90^\circ]$ is a limit angle adjusted to the TF. 

Having $\alpha$ for the sigmoid, we determine its random orientation (rotation) selecting randomly from any symmetric interval $n$ first components of the normal vector $\mathbf{n}$, and calculating the last one from \eqref{eqa0}. Repeating this for each sigmoid from the CSs set, we achieve a set of CSs randomly rotated around the y-axis, having a random slope angle greater than $\alpha_{min}$. Finally, we distribute the CSs into the input space. To do so, for each $i$-th sigmoid we randomly select point $\mathbf{x}_i^*$ from the input space and shift the sigmoid in such a way that its inflection point is in $\mathbf{x}_i^*$. Thus: 

%wprowadzić $\mathbf{x}^*_i$

\begin{equation}
\frac{1}{1 + \exp(-(a_{i,1}x_{i,1}^* + ...+a_{i,n}x_{i,n}^*+b_i))}=0.5
\label{eqSig4}
\end{equation}
Hence:

\begin{equation}
b_i=-a_{i,1}x_{i,1}^* - ...-a_{i,n}x_{i,n}^*
\label{eqbi2}
\end{equation}

Note that the biases of the hidden nodes are not selected from a certain interval, as they are in the typical approach, but they are dependent on the node weights. Equation \eqref{eqbi2} ensures the distribution of CSs in the input space according to the data distribution.

\subsection{Discussion}

The proposed method of generating random parameters of hidden nodes ensures that the steep fragments of CSs are put inside the input space. The input space is an $ n $-dimensional hypercube $ H = [x_{\min 1}, x_{\max 1}]\times[x_{\min 2}, x_{\max 2}]\times ... \times[x_{\min n}, x_{\max n}] $, where $x_{\min k}$ and $x_{\max k}$ are the lower and upper bounds, respectively,  for data in the dimension $k$. In most cases it is convenient to normalize all input variables into the range $[0,1]$, so the input hypercube is $ H = [0, 1]^n \subset \mathbb{R}^n $.

The steepness of CSs is controlled by the slope angle $\alpha_{min}$, i.e. the CSs generated have a slope angle from $\alpha_{min}$ to $90^\circ$. If we want to limit the maximum value of the slope angle, we can use $\alpha_{max}>\alpha_{min}$ as an upper bound for $\alpha$:      
\begin{equation}
\Delta = (\alpha_{min}, \alpha_{max})
\label{eqInt2}
\end{equation}

Both parameters $\alpha_{min}$ and $\alpha_{max}$ should be adjusted to TF, e.g. in the cross-validation, such as the number of hidden nodes $m$. The number of hidden nodes depends on the TF as well. More complex TFs need more neurons to get sufficient approximation accuracy. To minimize the computational effort in the hyperparameters selection phase, we can consider fixed $\alpha_{min}=0^\circ$ and $\alpha_{max}=90^\circ$ and only search for $m$. If this does not bring satisfactory results we can also search for $\alpha_{min}$ assuming fixed $\alpha_{max}=90^\circ$. Finally, when this is still unsatisfactory we can also search for $\alpha_{max}$.

The CS rotation is determined by the normal vector to the tangent hyperplane $T$. Its first $n$ components are selected randomly, independently and uniformly from the same interval, which should be symmetrical, i.e. $[-d,d]$. In such a case each rotation is just as likely. The interval limit value $d$ is not important. We recommend the interval for $a_k^{\prime}$ selection as $[-1, 1]$. The last component of the normal vector, $a_0^{\prime}$, is calculated from \eqref{eqa0} to ensure the slope of the CSs at an angle of $\alpha$. The sign for $a_0^{\prime}$ is selected randomly. 

In the bias determination step we select points $\mathbf{x}^*$ from hypercube $H$ and then shift the sigmoids to these points. To avoid shifting CSs to the empty region without data, it is reasonable to shift them to the points from the training set. In such a case, points $\mathbf{x}^*$ are the randomly selected training points. This ensures that all CSs have their steep fragments in the regions containing data. Another way of choosing points $\mathbf{x}^*$ is to group training points into $m$ clusters and take the prototypes $\mathbf{p}$ of these clusters (centroids) as $\mathbf{x}^*$. 

In this work, sigmoids are considered as activation functions of the hidden nodes. But similar considerations can be made for other types of activation functions, such as Gaussian, softplus, sine and others (see \cite{Dud19}, where an alternative method of random parameters generation was proposed and different activation functions were considered).

The procedure of generating the random parameters of FNNs described above is summarized in Algorithm 1. 

\section{Simulation Study}
The proposed method of selecting random parameters for FNN is illustrated by several examples. The first one concerns a single-variable function approximation, where TF is in the form:
\begin{equation}
g(x) = 0.2e^{-\left(10x - 4\right)^2} + 0.5e^{-\left(80x - 40\right)^2} + 0.3e^{-\left(80x - 20\right)^2}
\label{eqX}
\end{equation} 

\begin{algorithm}[H]
	\caption{Generating Random Parameters of FNNs}
	\label{alg1}
	\begin{algorithmic}
		\STATE {\bfseries Input:}\\ 
		\vspace{4mm}
		Number of hidden nodes $m$\\
		Number of inputs $n$\\
		Input hypercube $H$\\
		Training set $\Phi$ (optionally)\\
		Set of prototypes $\{\mathbf{p}_i\}_{i = 1, ..., m}$ (optionally)\\
		\vspace{4mm}
		\STATE {\bfseries Output:}\\ 
		\vspace{4mm}
		Weights $ \mathbf{A} = \left[
		\begin{array}{ccc}
		a_{1,1} & \ldots & a_{m,1} \\
		\vdots & \ddots & \vdots \\
		a_{1,n} & \ldots & a_{m,n}
		\end{array}
		\right]	$  \\
		Biases $ \mathbf{b} = [b_1, \ldots, b_m] $ \\    
		\vspace{4mm}
		\STATE {\bfseries Procedure:}\\
		\vspace{4mm}
		Set $\alpha_{min}=0^\circ$ or choose $\alpha_{min}\in[0^\circ, 90^\circ]$\\
		Set $\alpha_{max}=90^\circ$ or choose $\alpha_{max}\in[\alpha_{min}, 90^\circ]$\\
		\FOR{$i=1$ {\bfseries to} $m$}
		
		\STATE Choose randomly $\alpha \sim U(\alpha_{min}, \alpha_{max})$
		\STATE Choose randomly i.i.d. $ a_1^{\prime}, \ldots, a_n^{\prime} \sim U(-1, 1) $ 
		\STATE Calculate 		
		\begin{equation*}
		a_0^{\prime} = (-1)^c \frac{\sqrt{(a_1^{\prime})^2+...+(a_n^{\prime})^2}}{\tan \alpha}
		\end{equation*}
		\hspace{4mm} where $c \sim U\{0, 1\}$ \\ 
		\STATE Calculate 
		\begin{equation*}
		a_{i,k} = -4\frac{a_k^{\prime}}{a_0^{\prime}} \quad \text{for } k = 1, 2, ..., n  
		\end{equation*}
		\STATE Choose randomly $\mathbf{x}^*=[x_1^*, ..., x_n^*]$ from $H$\\
		\hspace{4mm}or set $\mathbf{x}^* = \mathbf{x}_j \in \Phi $, 
		where $j \sim U\{1, 2, \ldots, N\}$ \\
		\hspace{4mm}or set $\mathbf{x}^* = \mathbf{p}_i $, where $ \mathbf{p}_i $ is a~prototype of the $ i $-th \\
		\hspace{8mm}cluster of $ \mathbf{x} \in \Phi $
		\STATE Calculate 
		\begin{equation*}
		b_i = -\sum\limits_{k=1}^n a_{i,k}x_k^*
		\end{equation*}
		\ENDFOR
	\end{algorithmic}
\end{algorithm}

The training set contains $ 1000 $ points $ (x_l, y_l) $, where $ x_l $ are uniformly randomly distributed on $ [0, 1] $ and $ y_l $ are calculated from $\eqref{eqX}$. A test set of size $ 300 $ is generated from a~regularly spaced grid on $ [0, 1] $.

\begin{figure}
	\centering
	\includegraphics[width=0.49\textwidth]{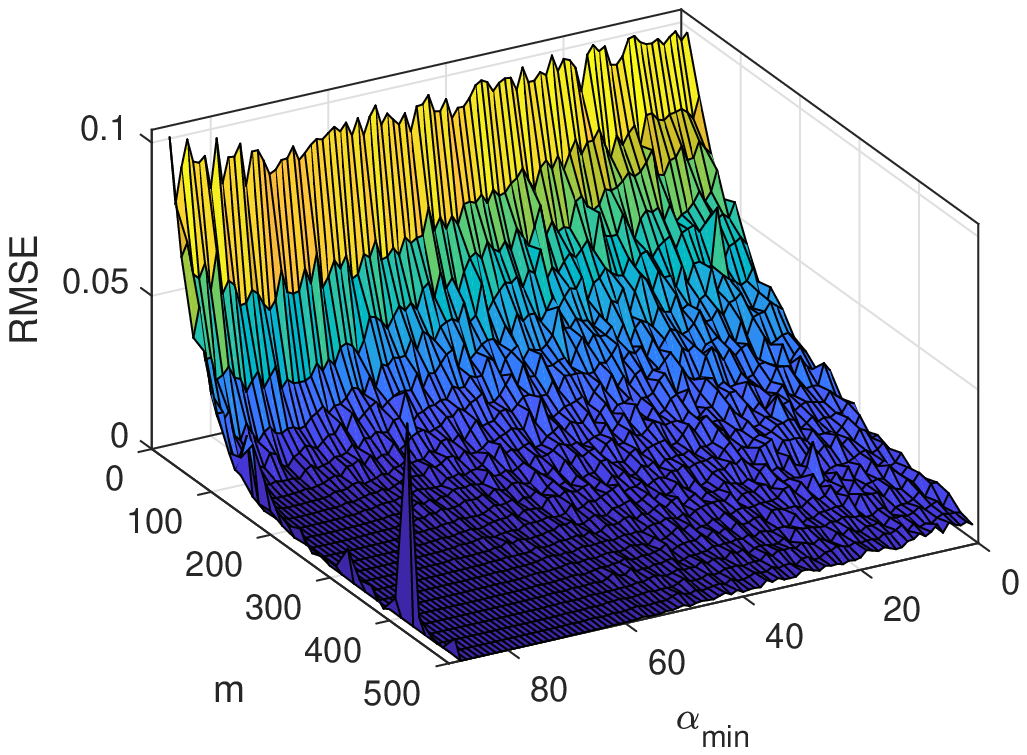}
	\includegraphics[width=0.49\textwidth]{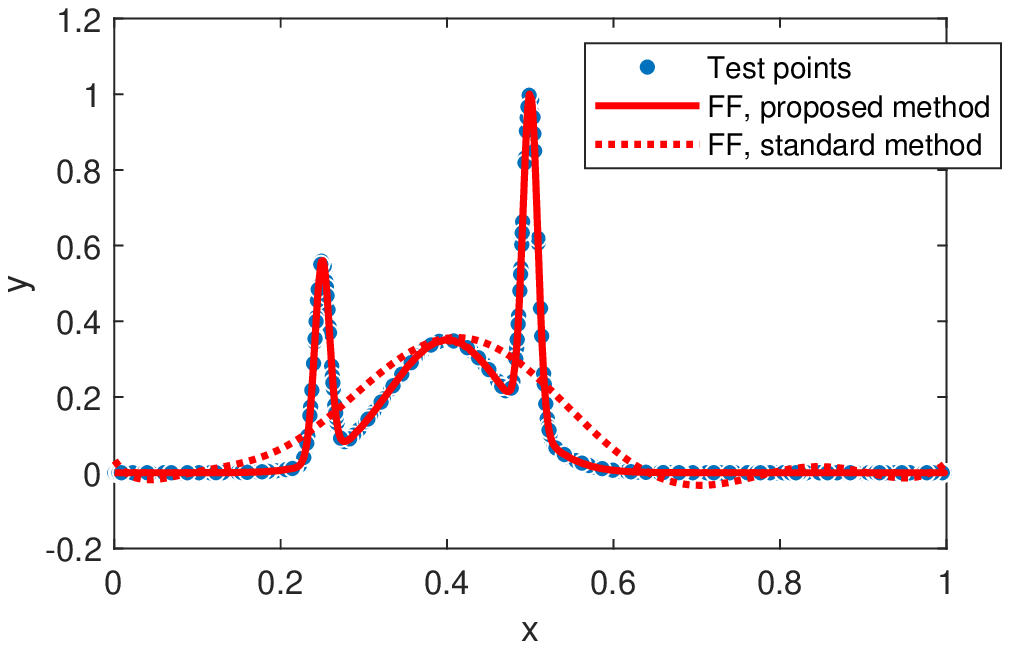}
	\caption{TF $\eqref{eqX}$ fitting using the proposed method: RMSE in the grid search (left panel) and fitted curves (right panel).}
	\label{Fig4}
\end{figure}

The left panel of Fig. \ref{Fig4} shows root-mean-square error (RMSE) while searching for the number of hidden nodes $m$ and $\alpha_{min}$ in 10-fold cross-validation procedure ($\alpha_{max}$ was set as $90^\circ$). The lowest RMSE ($1.34\cdot 10^{-6}$) was achieved for $\alpha_{min}=85$ and $m=320$. The RMSE on the test set was $9.35\cdot 10^{-7}$ at the optimal values of the hyperparameters. When using the standard method of random parameters generation, i.e. the weights and biases selected from the interval $[-1, 1]$, the test RMSE was above $0.1$ regardless of the number of neurons. FFs for the proposed and standard methods of random parameters generation are shown in the right panel of Fig. \ref{Fig4} ($500$ nodes were used for the standard method). Note that the standard method is not able to fit the TF. This is because CSs are too flat in the II.

In the second example we use two-variable TF in the form: 

\begin{equation}
g(\mathbf{x}) = \sin\left(20 e^{x_1}\right)\cdot x_1^2 + \sin\left(20e^{x_2}\right)\cdot x_2^2
\label{eqG}
\end{equation}

\begin{figure}
	\centering
	\includegraphics[width=0.49\textwidth]{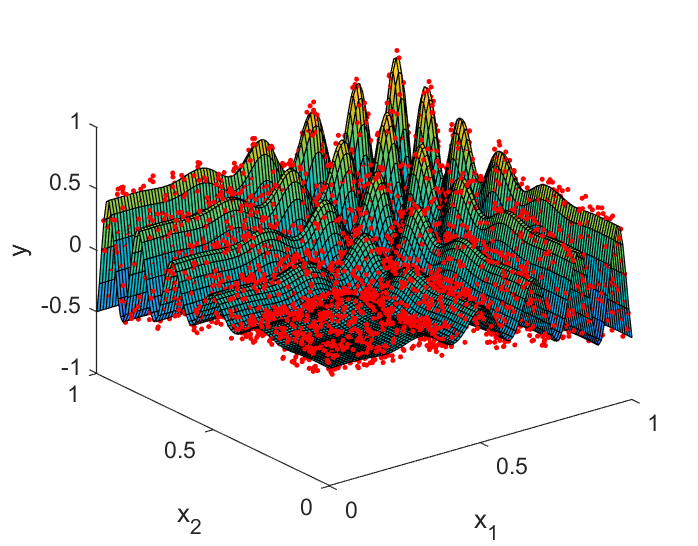}
	\caption{TF $\eqref{eqG}$ and the training points.}
	\label{Fig6}
\end{figure}

The training set contains $5000$ points $(\mathbf{x}_l, y_l)$, where both components of $\mathbf{x}_l$ are independently uniformly randomly distributed on $[0, 1]$ and $y_l$ are distorted by adding the uniform noise distributed in $[–0.2, 0.2]$. The test set containing $100000$ points is distributed in the input space on a~regularly spaced grid and is not disturbed by noise.The outputs are normalized in the range $[–1, 1]$. The TF and training points are shown in Fig. \ref{Fig6}. Note that variation of the TF is the lowest in the corner $(0, 0)$ and increases towards the corner $(1, 1)$.

\begin{figure}
	\centering
	\includegraphics[width=0.49\textwidth]{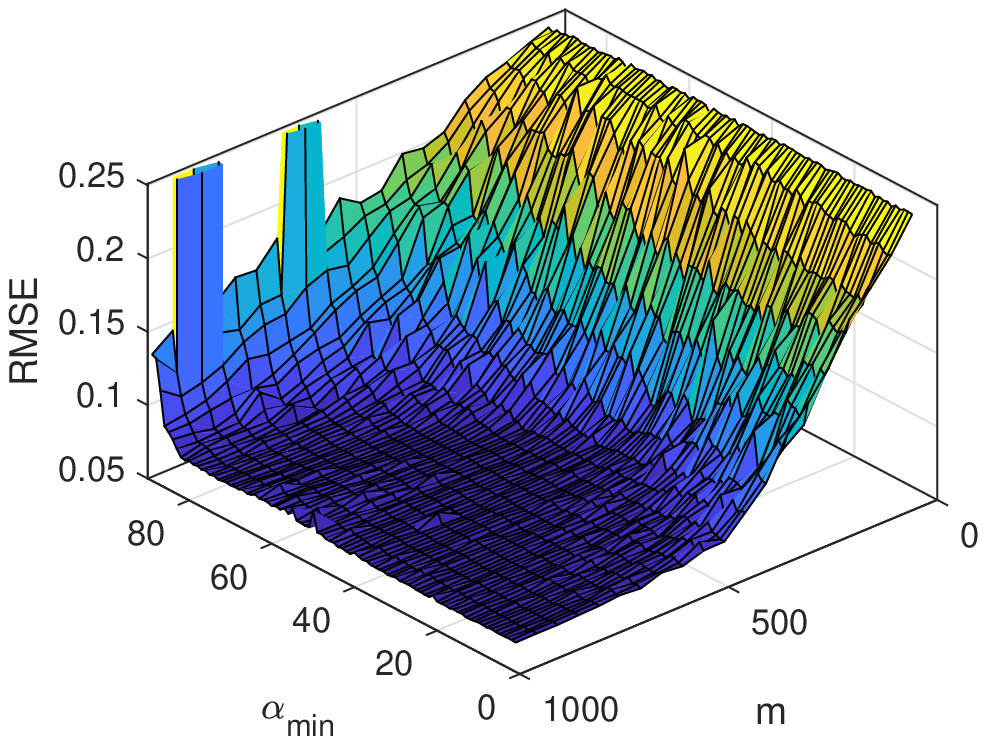}
	\includegraphics[width=0.49\textwidth]{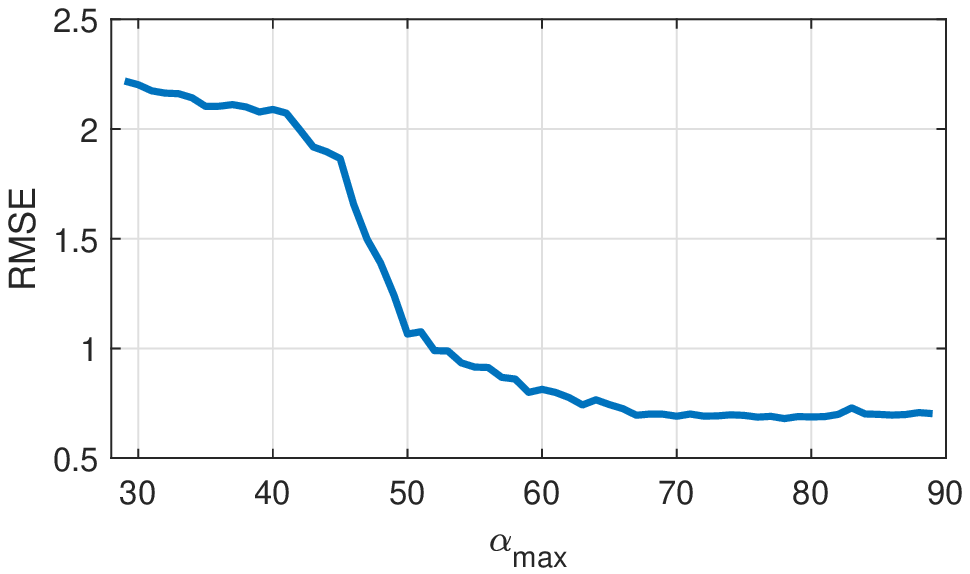}
	\caption{TF $\eqref{eqG}$ fitting using the proposed method: RMSE in the grid search (left panel) and impact of $\alpha_{max}$ on RMSE at $\alpha_{min}=29^\circ$ and $m=700$ (right panel).}
	\label{Fig7}
\end{figure}

\begin{figure}
	\centering
	\includegraphics[width=0.49\textwidth]{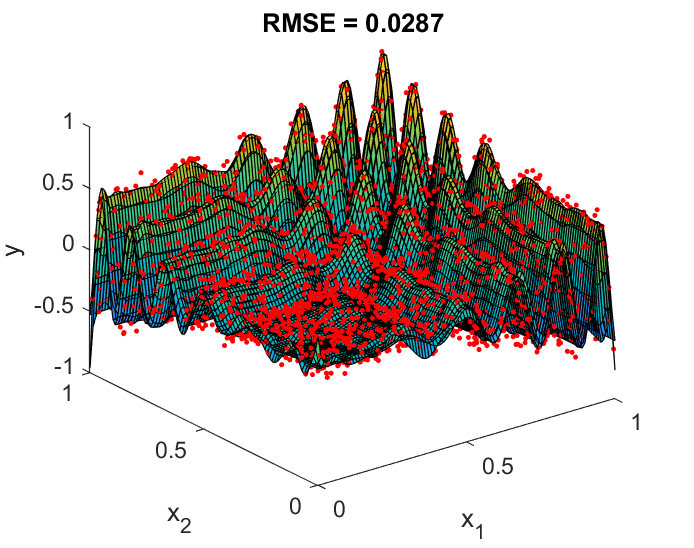}
	\includegraphics[width=0.49\textwidth]{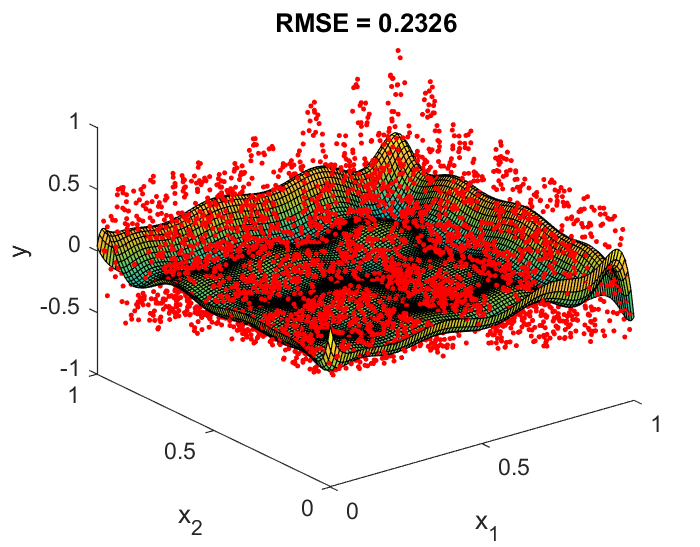}
	\caption{Fitted surface for TF $\eqref{eqG}$, the proposed (left panel) and standard method (right panel).}
	\label{Fig8}
\end{figure}  

The left panel of Fig. \ref{Fig7} shows RMSE while searching for the hyperparameters values in a 10-fold cross-validation procedure. In this figure, we can observe a large plateau region for $0^\circ<\alpha_{min}<70^\circ$ and $m>500$. The RMSE in this region is less than 0.0850 at the lowest value of 0.0690 for $\alpha_{min}=29^\circ$ and $m=700$. 

The right panel of Fig. \ref{Fig7} shows RMSE when changing the upper bound slope angle $\alpha_{max}$ from  $\alpha_{min}$ to $90^\circ$ at the optimal values for $m$ and $\alpha_{min}$. As we can see from this figure, the lowest error is when $\alpha_{max}$ is above $70^\circ$.

Fig. \ref{Fig8} shows the FF when using the proposed method with the optimal values of hyperparameters and the standard method when $700$ neurons are used (increasing the neuron number did not improve the results). As we can see from these figures, the proposed method maps the TF quite well (RMSE = $0.0287$) while the standard method fails (RMSE = $0.2326$). 

The last example concerns a 21-dimensional modeling problem: Compactiv - the Computer Activity dataset  which is a collection of computer systems activity measures. The data was collected from a Sun Sparcstation 20/712 with 128 Mbytes of memory running in a multi-user university department. The task is to predict the portion of time that CPUs run in user mode. There are 8192 samples composed of 21 input variables (activity measures) and one output variable. The whole data set was divided into a training set containing $75\%$ of samples selected randomly, and a test set containing the remaining samples. The dataset was downloaded from KEEL (Knowledge Extraction based on Evolutionary Learning) dataset repository (http://www.keel.es). The input and output variables are normalized into $[0, 1]$. 

\begin{figure}
	\centering
	\includegraphics[width=0.49\textwidth]{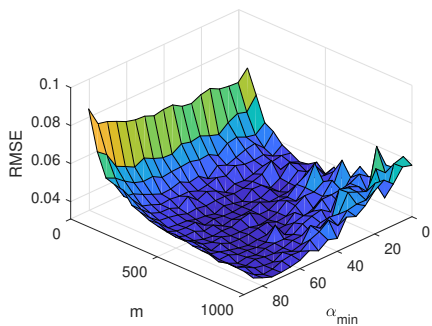}
	\caption{Compactive data fitting using the proposed method: RMSE in the grid search.}
	\label{Fig10}
\end{figure}

The RMSE in the grid search procedure using a 10-fold cross-validation in Fig. \ref{Fig10} is shown. The lowest value of RMSE was $0.0309$ for $\alpha_{min}=45^\circ$ and $m=600$. The mean value of the test error for $100$ trials of the learning sessions carried out at the optimal values of hyperparameters was $0.0335$. For the standard method it was $0.0358$. The difference between RMSE for the proposed and standard method is not as high as for the TFs \eqref{eqX} and \eqref{eqG}. This is probably because the TF in this case has no strong fluctuations and can be modeled using flat neurons.   

\section{Conclusions}
\label{con}

One of the most important issues in the randomized learning of FNNs is the selection of random parameters: weights and biases of hidden nodes. In the existing methods the random parameters are selected uniformly from the fixed interval, such as $[-1, 1]$ or another symmetrical interval whose bounds are adjusted to the problem being solved. We have shown that the weights and biases of the hidden nodes have a different meaning and should not be selected from the same interval. 

In this work we recommend generating random parameters in single-hidden-layer FNNs taking into account the input space location and size, target function complexity, and activation function type. We propose a method which randomly selects the slope angles for the activation functions from the interval adjusted to the target function. Then, after rotating randomly the activation functions and shifting them into the input hypercube, we calculate weights and biases. The proposed approach turned out to be much more accurate than the existing one in regression problems. In the simulation study generating random parameters from the fixed interval $[-1, 1]$ brought very poor fitting, while the proposed method performed very well on target functions with strong fluctuations.

Future work will focus on the better adjustment of the hidden nodes to data. This should translate into a more compact network structure without redundant nodes. Additionally, the adaptation of the method to classification problems is planned.


\end{document}